\documentclass[runningheads]{llncs}
\usepackage{graphicx}
\usepackage{hyperref}
\begin{document}
\title{L3Cube-HindBERT and DevBERT: Pre-Trained BERT Transformer models for Devanagari based Hindi and Marathi Languages\thanks{Supported by L3Cube Pune.}}
\titlerunning{L3Cube-HindBERT and DevBERT}
%
\author{Raviraj Joshi}
\authorrunning{R. Joshi}
%
\institute{Indian Institute of Technology Madras \and
L3Cube Pune \\
\email{ravirajoshi@gmail.com}}
\maketitle              
\begin{abstract}
The monolingual Hindi BERT models currently available on the model hub do not perform better than the multi-lingual models on downstream tasks. We present L3Cube-HindBERT, a Hindi BERT model pre-trained on Hindi monolingual corpus. 
Further, since Indic languages, Hindi and Marathi share the Devanagari script, we train a single model for both languages. We release DevBERT, a Devanagari BERT model trained on both Marathi and Hindi monolingual datasets. We evaluate these models on downstream Hindi and Marathi text classification and named entity recognition tasks. The HindBERT and DevBERT-based models show significant improvements over multi-lingual MuRIL, IndicBERT, and XLM-R. Based on these observations we also release monolingual BERT models for other Indic languages Kannada, Telugu, Malayalam, Tamil, Gujarati, Assamese, Odia, Bengali, and Punjabi. 
These models are shared at \url{https://huggingface.co/l3cube-pune} .

\keywords{Hindi BERT  \and Marathi BERT \and Devanagari BERT \and Indic BERT \and Transformers \and Natural Language Processing \and Indian Low Resource Languages.}
\end{abstract}

\section{Introduction}
Pre-trained masked language models have become the de-facto standard for natural language processing tasks \cite{qiu2020pre}. These are Transformer based language models pretrained on a large amount of unsupervised data \cite{devlin2018bert,liu2019roberta}. The pre-trained models have also been useful for low-resource languages where the amount of annotated data is less \cite{kakwani2020indicnlpsuite,khanuja2021muril}. It is also common to train multi-lingual models so that related low-resource languages can benefit from each other. While multi-lingual BERT models perform well for extremely low-resource languages, it is shown that monolingual counterparts work better if we have sufficient monolingual data \cite{joshi2022l3cube}. 

In this work, we consider low-resource Indian languages Hindi and Marathi \cite{joshi2019deep,kulkarni2022experimental}. The monolingual Hindi BERT models available on the model hub do not perform competitively with multi-lingual models \cite{litake2022mono,jain2020indic}. This is despite having sufficient monolingual corpus in the target language. However, for Marathi\footnote{\url{https://github.com/l3cube-pune/MarathiNLP}} \cite{joshi2022l3cubeMahaNLP}, existing monolingual models based on L3Cube-MahaBERT are shown to work well than their multi-lingual counterparts \cite{joshi2022l3cube,velankar2022mono,litake2022mono,patil2022l3cube,velankar2022l3cube}. We, therefore, release a similar set of models for Hindi termed L3Cube-HindBERT pre-trained on publicly available Hindi monolingual datasets. These models are trained on roughly 1.8 B Hindi tokens.

Both Hindi and Marathi share the Devanagari script, so we further train BERT models on Hindi and Marathi monolingual data. We wish to exploit the similarities between closely related languages Hindi and Marathi and train a single model for both languages. We term these sets of models as L3Cube-DevBERT, the pre-trained transformer models trained on Devanagari data. These models are trained on roughly 2.5 B Devanagari tokens. 

\section{Related Work}
Recently, monolingual BERT models have been released for some Indian languages. The BERT models for Marathi termed as MahaBERT were released by L3Cube \cite{joshi2022l3cube}. They also released MahaRoBERTa and MahaAlBERT, Marathi models based on RoBERTa and AlBERT architecture respectively. These models were shown to provide state-of-the-art performance on text classification and named entity recognition tasks. Another pre-trained language model from L3Cube includes HingBERT \cite{nayak-joshi-2022-l3cube}, which was pre-trained on Hindi-English code mixed tweets. The model was evaluated on GLUECoS and provided a superior performance on the majority of the tasks. Further monolingual BERT models for Hindi, Bengali, and Telegu were released in \cite{jain2020indic}. The models for Hindi and Bengali showed state-of-the-art performance on the respective text classification tasks. However, they notice a marginal improvement in performance with monolingual BERT models as compared to the multi-lingual models. Similarly, monolingual models for the Bengali language were also released in \cite{bhattacharjee2022banglabert} and \cite{kowsher2022bangla}. These models were trained on around 27.5GB and 40GB of Bengali corpus respectively. Dedicated BERT models for Telugu were released in \cite{marreddy2022resource}. These models were trained on a Telugu corpus consisting of around 8 million sentences and were shown to provide superior results on multiple tasks. Therefore, documented monolingual BERT models are mainly available for Marathi, Bengali, and Telugu. These models are still missing for major Indian languages like Hindi, Kannada, Malayalam, Gujarati, etc. The available options for these languages include multi-lingual models like mBERT \cite{devlin2018bert}, xlm-RoBERTa \cite{conneau2019unsupervised}, IndicBERT \cite{kakwani2020indicnlpsuite}, and MuRIL \cite{khanuja2021muril}.

\section{Experimental Setup}
We fine-tune the existing multilingual models like MuRIL \cite{khanuja2021muril}, xlmRoBERTa \cite{conneau2020unsupervised}, and IndicBERT \cite{kakwani2020indicnlpsuite} on the monolingual corpus. The respective Hindi models are termed as HindBERT\footnote{\url{https://huggingface.co/l3cube-pune/hindi-bert-v2}}, HindRoBERTa\footnote{\url{https://huggingface.co/l3cube-pune/hindi-roberta}}, and HindAlBERT\footnote{\url{https://huggingface.co/l3cube-pune/hindi-albert}}. The corresponding Devanagari models are termed as DevBERT\footnote{\url{https://huggingface.co/l3cube-pune/hindi-marathi-dev-bert}}, DevRoBERTa\footnote{\url{https://huggingface.co/l3cube-pune/hindi-marathi-dev-roberta}}, and DevAlBERT\footnote{\url{https://huggingface.co/l3cube-pune/hindi-marathi-dev-albert}}. We also train MahaBERT-scratch\footnote{\url{https://huggingface.co/l3cube-pune/marathi-bert-scratch}}, HindBERT-scratch\footnote{\url{https://huggingface.co/l3cube-pune/hindi-bert-scratch}} and DevBERT-scratch\footnote{\url{https://huggingface.co/l3cube-pune/hindi-marathi-dev-bert-scratch}} from scratch with custom trained tokenizer. However, we evaluate only fine-tuned models and the evaluation of models trained from scratch is left to future scope. These models are shared publicly on HuggingFace \cite{wolf2019huggingface} model hub. The hyper-parameter setup used for training and fine-tuning these models is the same as described in \cite{joshi2022l3cube}. 

\section{Downstream Datasets}
We evaluate HindBERT and DevBERT-based models on downstream text classification and named entity recognition tasks in Hindi and Marathi. We show the superior performance of models for two major Indian languages. 
The downstream evaluation is performed on the following datasets.
\begin{itemize}
    \item \textbf{IITP Movie Reviews (Hindi)} \cite{akhtar2016hybrid}: A Hindi movie reviews dataset with individual samples categorized into positive, negative, and neutral classes. The dataset consists of 2479 train, 309 validation, and 309 test samples.
    \item \textbf{WikiAnn Hindi NER dataset} \cite{pan2017cross}: A Hindi named entity recognition dataset consisting of 11833 sentences. The tokens in each of the sentences have been categorized into three classes namely Organization, Person, and Location. 
    \item \textbf{L3Cube-MahaSent (Marathi)} \cite{kulkarni2021l3cubemahasent}: A tweet-based sentiment analysis dataset in Marathi. The target labels are positive, negative, and neutral. The dataset consists of 2114 train, 2250 test, and 1500 validation samples. 
    \item \textbf{L3Cube-MahaNER (Marathi)} \cite{patil2022l3cube}: It is a Marathi named entity recognition corpus from the news domain. The dataset consists of 25k sentences with individual tokens categorized into 8 named entities. 
\end{itemize}


    \begin{table*}
\begin{center}
\begin{tabular}{{p{2.5cm}p{2cm}p{2cm}p{2cm}p{2cm}}}
\hline \textbf{Model} & \textbf{IITP-Movies} & \textbf{WikiAnn NER} & \textbf{L3Cube-MahaSent} & \textbf{L3Cube-MahaNER} \\
\hline
MuRIL & 66.00 & 84.9 & 84.3 & 87.82\\
indicBERT &  65.00 & 82.65 & 83.3 & 86.56\\
XLM-R & 67.00 & 83.04 & 82.0 & 85.69\\ \hline
HindBERT &  \textbf{69.00} & \textbf{86.00} & - & -\\ 
HindAlBERT & 65.00 & 83.46 & - & -\\ 
HindRoBERTa & 67.00 & 84.00 & - & -\\ \hline
MahaBERT &  - & - & 84.9 & 88.0 \\  \hline
DevBERT &  67.00 & 85.0 & \textbf{85.00} & \textbf{88.31}\\ 
DevAlBERT & 65.00 & 82.46 & 84.4 & 87.50\\ 
DevRoBERTa & 67.00 & 83.14 & 83.2 & 88.25\\ \hline

\hline
\end{tabular}
\caption{\label{result-tab} The results for different models on classification and NER tasks. The numbers for classification task IITP-Movies and L3Cube-MahaSent represent the classification accuracy. The numbers for the NER task WikiAnn Hindi NER and L3Cube-MahaNER represent the macro-f1 score.}
\end{center}
\end{table*}

\section{Results and Discussion}
The HindBERT and DevBERT models are evaluated on classification and NER tasks. The results are described in Table \ref{result-tab}. We show that these monolingual models are competitive with their multilingual counterparts and provide superior performance in all of the downstream tasks. 

The HindBERT models outperform both multi-lingual models and DevBERT models on Hindi datasets. The DevBERT models are still desirable as a single model works competitively for both Hindi and Marathi. Moreover, DevBERT marginally outperforms MahaBERT\footnote{\url{https://huggingface.co/l3cube-pune/marathi-bert-v2}}\cite{joshi2022l3cube} on Marathi tasks. The performance of DevBERT models is better than multi-lingual models on both Hindi and Marathi datasets. 

Based on these observations for Hindi and Marathi we also release monolingual models for other Indian languages. We also release BERT models for Indic languages Kannada, Telugu, Malayalam, Tamil, Gujarati, Assamese, Odia, Bengali, and Punjabi.
These models are termed as 
Kannada-BERT\footnote{\url{https://huggingface.co/l3cube-pune/kannada-bert}}, 
Telugu-BERT\footnote{\url{https://huggingface.co/l3cube-pune/telugu-bert}}, 
Malayalam-BERT\footnote{\url{https://huggingface.co/l3cube-pune/malayalam-bert}}, 
Tamil-BERT\footnote{\url{https://huggingface.co/l3cube-pune/tamil-bert}}, 
Gujarati-BERT\footnote{\url{https://huggingface.co/l3cube-pune/gujarati-bert}}, 
Assamese-BERT\footnote{\url{https://huggingface.co/l3cube-pune/assamese-bert}}, 
Odia-BERT\footnote{\url{https://huggingface.co/l3cube-pune/odia-bert}}, 
Bengali-BERT\footnote{\url{https://huggingface.co/l3cube-pune/bengali-bert}}, and 
Punjabi-BERT\footnote{\url{https://huggingface.co/l3cube-pune/punjabi-bert}}.
We further release Kannada-BERT-Scratch \footnote{\url{https://huggingface.co/l3cube-pune/kannada-bert-scratch}}, 
Telugu-BERT-Scratch\footnote{\url{https://huggingface.co/l3cube-pune/telugu-bert-scratch}}, 
Malayalam-BERT-Scratch\footnote{\url{https://huggingface.co/l3cube-pune/malayalam-bert-scratch}},
and Gujarati-BERT-Scratch\footnote{\url{https://huggingface.co/l3cube-pune/gujarati-bert-scratch}}.
These models have been trained from scratch with a custom tokenizer. Again evaluation of these models is left to future scope.

\section{Conclusion}
We present L3Cube-HindBERT and DevBERT models trained on monolingual Hindi corpus and Hindi $+$ Marathi corpus respectively. We highlight the lack of competitive Hindi BERT models in the public domain and hence come up with these models. The HindBERT models have been evaluated on Hindi classification and Hindi NER datasets whereas the DevBERT is evaluated on both Hindi and Marathi downstream tasks. The newly introduced models perform better than multi-lingual models on these downstream tasks. We also release monolingual BERT models for 9 other Indic languages. An exhaustive evaluation of these models is left to future scope.

\section*{Acknowledgements}
We are extremely thankful to Parth Patil, Abhishek Velankar, and Group Optimize\_Prime for their valuable contributions.

\bibliographystyle{splncs04}
\bibliography{main}

\end{document}